\documentclass{article}
\usepackage{spconf,amsmath,epsfig}

\usepackage{graphicx}
\usepackage{bm}
\usepackage{enumerate}
\usepackage{amsfonts}
\usepackage{subfigure}
\usepackage{tabularx}
\usepackage[warn]{textcomp}
\usepackage[para,online,flushleft]{threeparttable}

\begin{document}\sloppy
\ninept 

\def\x{{\mathbf x}}
\def\L{{\cal L}}

\newcommand{\etal}{\textit{et al}.} 
\newcommand{\ie}{\textit{i}.\textit{e}.}
\newcommand{\eg}{\textit{e}.\textit{g}.}
\newcommand{\myfont}{\fontsize{8pt}{\baselineskip}\selectfont}

\title{Skeleton-Based Action Recognition with Synchronous Local and Non-local Spatio-temporal Learning and Frequency Attention}
%
\name{Guyue Hu\textsuperscript{1, 3}, Bo Cui\textsuperscript{1, 3}, Shan Yu\textsuperscript{1, 2, 3}}
\address{\textsuperscript{1}Brainnetome Center \& National Laboratory of Pattern Recognition, \\
	Institute of Automation, Chinese Academy of Sciences\\
	\textsuperscript{2}CAS Center for Excellence in Brain Science and Intelligence Technology\\
	\textsuperscript{3}University of Chinese Academy of Sciences\\
	\{guyue.hu, bo.cui, shan.yu\}@nlpr.ia.ac.cn}

\maketitle

\begin{abstract}
Benefiting from its succinctness and robustness, skeleton-based action recognition has recently attracted much attention. Most existing methods utilize local networks (\eg~recurrent, convolutional, and graph convolutional networks) to extract spatio-temporal dynamics hierarchically. As a consequence, the local and non-local dependencies, which contain more details and semantics respectively, are asynchronously captured in different level of layers. Moreover, existing methods are limited to the spatio-temporal domain and ignore information in the frequency domain. To better extract synchronous detailed and semantic information from multi-domains, we propose a residual frequency attention (rFA) block to focus on discriminative patterns in the frequency domain, and a synchronous local and non-local (SLnL) block to simultaneously capture the details and semantics in the spatio-temporal domain. Besides, a soft-margin focal loss (SMFL) is proposed to optimize the learning whole process, which automatically conducts data selection and encourages intrinsic margins in classifiers. Our approach significantly outperforms other state-of-the-art methods on several large-scale datasets.
\end{abstract}
\section{Introduction}
\label{sec:intro}
The skeleton-based human action recognition has recently attracted much attention due to its succinctness of representation and robustness to variations of viewpoints, appearances and surrounding distractions \cite{DBLP:conf/iccv/ZhangLXZXZ17}. Most previous works treat skeletal actions as sequences and pseudo-images, then apply Recurrent Neural Networks (RNN) \cite{DBLP:conf/iccv/ZhangLXZXZ17,DBLP:conf/cvpr/ShahroudyLNW16,DBLP:conf/eccv/LiuSXW16} and Convolutional Neural Networks (CNN) \cite{DBLP:conf/cvpr/KeBASB17,DBLP:conf/ijcai/LiZXP18} to model the temporal evolutions and the spatio-temporal dynamics, respectively. Yan \etal~\cite{DBLP:conf/aaai/YanXL18} also feeds skeleton graphs into graph convolutional networks (GCN) to exploit the structure information of human body. However, all the aforementioned methods apply stacked local networks to hierarchically extract spatio-temporal features, which exist two serious problems. 1) The recurrent and convolutional operations are neighborhood-based local operations \cite{DBLP:journals/corr/abs-1711-07971}, so the local-range detailed information and non-local semantic information mainly be captured asynchronously in the lower and higher layers respectively, which hinders the fusion of details and semantics in action dynamics. 2) Human actions such as \textit{shaking hands}, \textit{brushing teeth}, and \textit{clapping} have characteristic frequency patterns, but previous works are always limited to the spatio-temporal dynamics and ignore periodic patterns in the frequency domain.

\begin{figure*}[tbp]
	\centering
	\includegraphics[width=0.95\linewidth,height=0.26\linewidth]{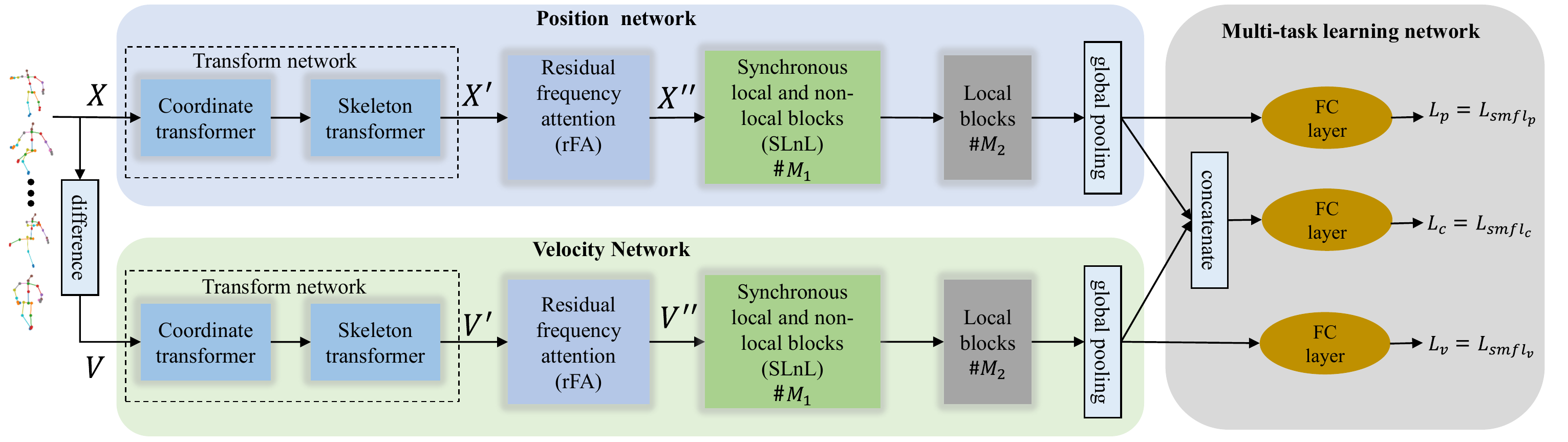}
	\caption{The overall pipeline of the proposed method. The position and velocity information of human joints are fed into a tranform network, a residual attention network, $ M_1 $ synchronous local and non-local blocks, and $ M_2 $ local blocks sequentially. Treated as a pseudo multi-task learning task, the proposed model is optimized according to our soft-margin focal loss.}
	\label{fig_pipline}
\end{figure*}

In this paper, we propose a novel model SLnL-rFA to better extract synchronous detailed and semantic information from multi-domains. SLnL-rFA is equipped with synchronous local and non-local (SLnL) blocks for spatio-temporal learning, and a residual frequency attention (rFA) block for frequency-patterns mining. To optimize whole learning process, a novel soft-margin focal loss (SMFL) is also proposed, which adaptively conducts data selection during training and encourages intrinsic margin in classifiers. Fig.\ref{fig_pipline} shows the 
pipeline of our method. Firstly, an adaptive transform network augments and transforms the skeletal actions. Secondly, the residual frequency attention block selects discriminative frequency patterns. Then, following with $ \it{M}_1 $ synchronous local and non-local (SLnL) blocks and $ \it{M}_2 $ local blocks in the spatio-temporal domain, where SLnL is designed to simultaneously extract local details and non-local semantics. Finally, three classifiers with inputs from position, velocity and concatenated features are optimized as a pseudo
multi-task learning problem according to our soft-margin focal loss.

Our main contributions are summarized as follows: 1) Moving beyond the spatio-temporal domain, we propose a residual frequency attention block to exploit frequency information for skeleton-based action recognition; 2) We propose a synchronous local and non-local block to simultaneously capture details and semantics in the early-stage layers; 3) We propose a soft-margin focal loss, which adaptively conducts data selection during training process and encourages intrinsic soft-margins in the classifiers; 
4) Our approach outperforms the state-of-the-art methods with significant margins on two large-scale datasets for skeleton-based action recognition.


\section{Related Works}
\textbf{Frequency domain analysis.}
Generalized frequency domain analysis contains several large classes of methods such as discret Fourier transform (DFT), short-time Fourier transform (SFT) and wavelet tranform, which are classical 
tools in the fields of signal analysis and image processing. 
Due to the booming of deep learning techniques \cite{DBLP:conf/nips/KrizhevskySH12,DBLP:conf/cvpr/HeZRS16}, 
methods based on the spatio-temporal domain dominate the field of computer vision, with only a few works paying attention to the frequency domain. For example, frequency domain analysis of critical points trajectories \cite{DBLP:conf/icip/BeaudryPM14} and frequency divergence image \cite{DBLP:conf/isbi/CruzS17} are applied for RGB-based action recognition. 
Our work will revisit the frequency domain, and exploit 
frequency patterns to improve the skeleton-based action recognition.

\textbf{Non-local operations.}
Non-local means is a classical filtering algorithm that allows distant pixels to contribute to the target pixel \cite{DBLP:conf/cvpr/BuadesCM05}. Block-matching \cite{DBLP:journals/tip/DabovFKE07} explores groups of non-local similarity between patches. 
Block-matching is widely used in computer vision tasks like super-resolution \cite{DBLP:conf/iccv/GlasnerBI09}, image inpainting \cite{DBLP:journals/tog/BarnesSFG09}, etc. 
The popular self-attention \cite{DBLP:conf/nips/VaswaniSPUJGKP17} in machine translation can also be viewed as a non-local operation. Recently, different non-local blocks are inserted into CNNs for video classification \cite{DBLP:journals/corr/abs-1711-07971} and RNNs for image restoration \cite{DBLP:journals/corr/abs-1806-02919}. However, their local and non-local operations apply to objects in different level of layers but our SLnL simultaneously operate on the same objects, thus only the proposed SLnL can extract local and non-local information synchronously.

\textbf{Reformed softmax loss.}
The softmax loss \cite{DBLP:conf/icml/LiuWYY16}, consisted of the last fully connected layer, the softmax function, and the cross-entropy loss, is widely applied in supervised learning due to its simplicity and clear probabilistic interpretation. However, recent works \cite{DBLP:conf/icml/LiuWYY16,DBLP:conf/ijcai/WangZLLGL18} 
have exposed its limitations on feature discriminability and have stimulated two types of methods for improvements. One type directly refines or combines the cross-entropy loss with other losses like contrastive loss, triplet loss, etc \cite{DBLP:conf/ijcai/WangZLLGL18,DBLP:conf/cvpr/SchroffKP15}. The other type reformulates the softmax function with geometrical or algebraic margin \cite{DBLP:conf/icml/LiuWYY16,DBLP:conf/ijcai/WangZLLGL18} to encourage intra-class compactness and inter-class separability of feature learning, which completely destroys the probabilistic meaning of the original softmax function. Our SMFL not only conducts data selection but also encourages intrinsic soft-margins in classifiers with a clear probabilistic interpretation.

\section{Methods}

\subsection{Preliminary}
A skeletal action {\myfont$\bm{X} \in \mathbb{R}^{d \times T\times N}$} is represented by $ \it{d} $ dimensional locations of {\myfont$ \it{N} $} body joints in a {\myfont$ \it{T} $} frame video. 
Following Li \etal~\cite{DBLP:conf/icmcs/LiZXP17}, we introduce a skeleton transformer to augment the number of joints and rearrange the order of joints. Similarly, a coordinate transformer is also applied to transform the original representations in single \textit{rectangular coordinate system} to rich representations in {\myfont$ \it{K} $} \textit{oblique coordinate systems}. The whole transform network in Fig.\ref{fig_pipline} is implemented with two fully connected layers and corresponding transpose, flatten, and concatenate operations. As a result, a new adaptive expression {\myfont$\bm{X'} \in \mathbb{R}^{Kd \times T'\times N'}$} is formed for each action. 

\begin{figure}[htb]
	\centering
	\includegraphics[width=0.9\linewidth,height=0.75\linewidth]{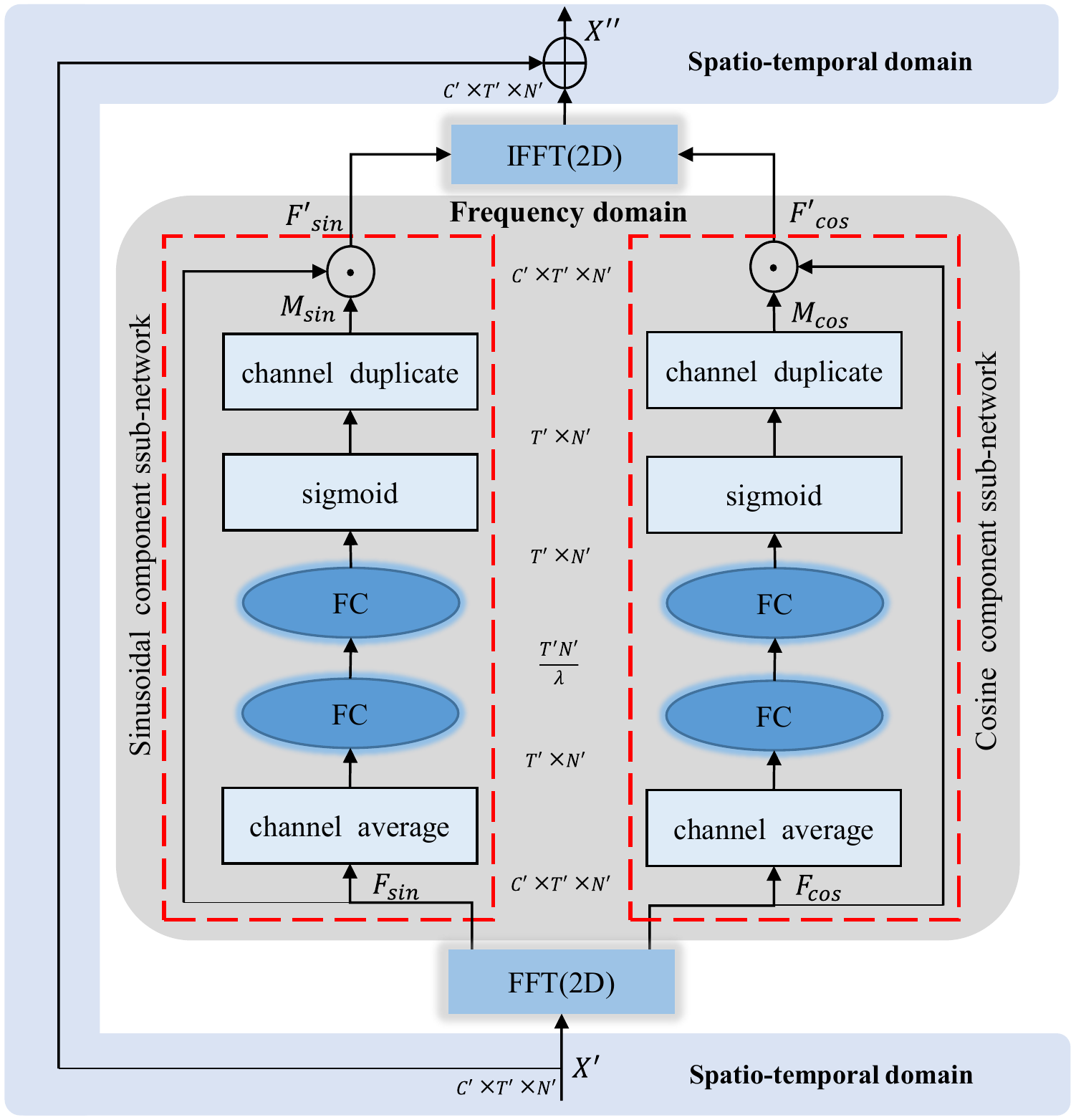}
	\caption{The residual frequency attention. The spatio-temparal domain and frequency domain are switched conveniently through 2D-FFT and 2D-IFFT. The attention for the sinusoidal and cosine components ($ \bm{F}_{sin} $, $ \bm{F}_{cos} $) are conducted in the frequency domain, and the residual component is applied in the spatio-temporal domain.}
	\label{fig2}
\end{figure}

\subsection{Residual Frequency Attention}
Previous works always concentrate on the spatio-temporal domain, but many actions contain inherent frequency-sensitive patterns, such as \textit{shaking hands}, and \textit{brushing teeth}, which motivates us to revisit the frequency domain. The classical operations in the frequency domain, such as high-pass, low-pass, and band-pass filters, only have a few parameters that are far from enough, thus we propose a more general frequency attention block (Fig. \ref{fig2}) equipped with abundant learnable parameters to adaptively select frequency components.

\begin{figure*}[htb]
	\centering
	\subfigure[2D Non-local module]{
		\label{Fig3:a}
		\includegraphics[width=0.21\linewidth,height=0.22\linewidth]{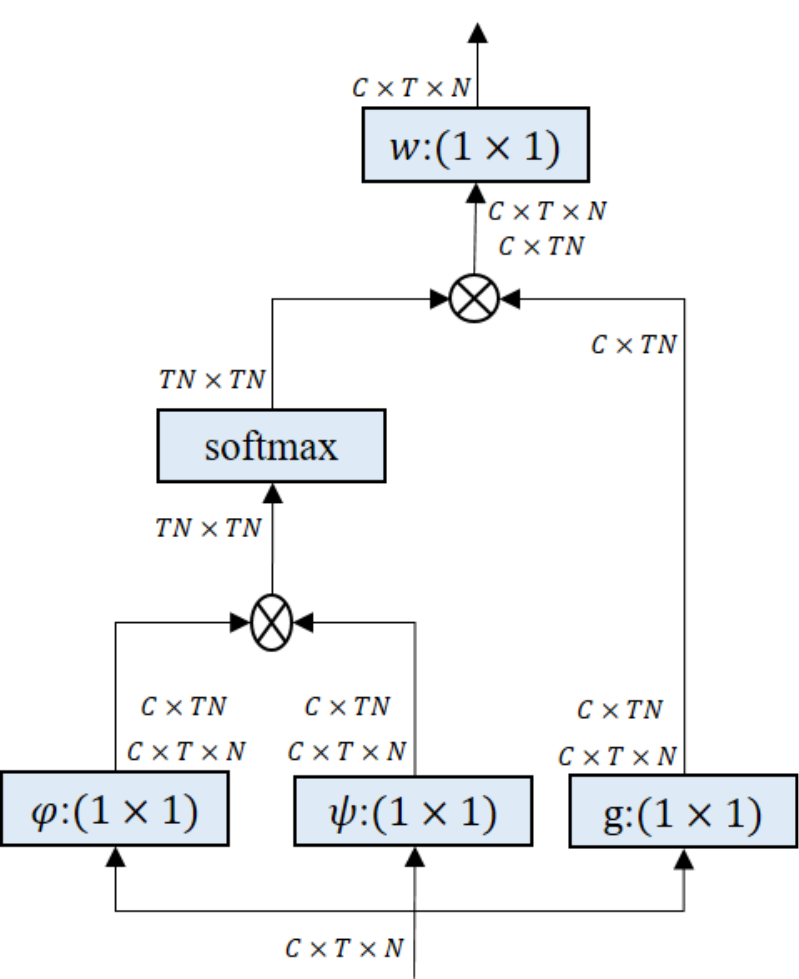}
	}
	\subfigure[Baseline local block]{
		\label{Fig3:b}
		\includegraphics[width=0.19\linewidth,height=0.22\linewidth]{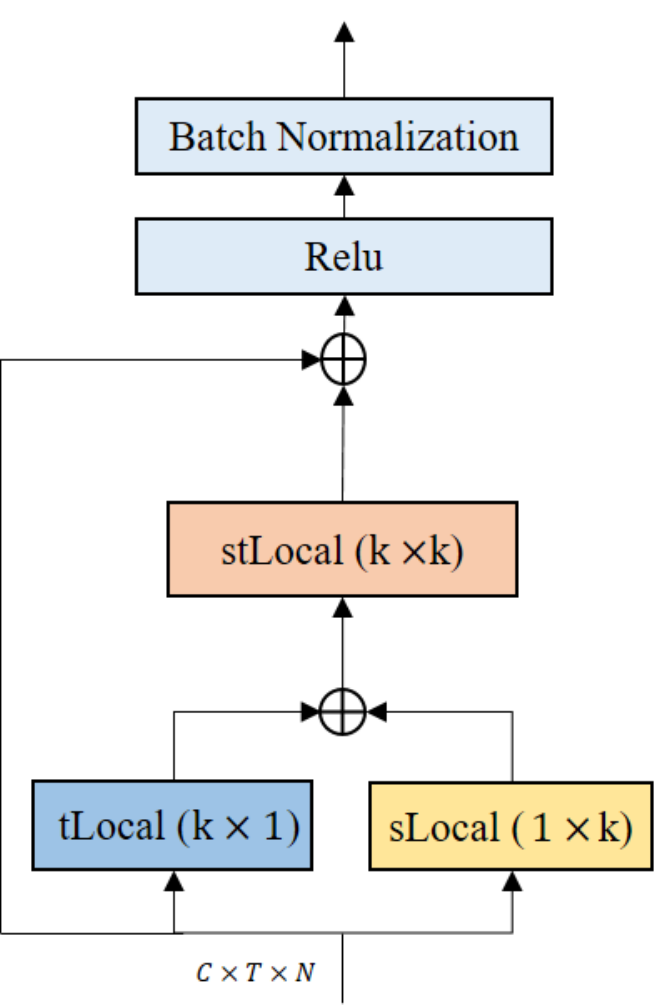}
	}
	\subfigure[SLnL block]{
		\label{Fig3:c}
		\includegraphics[width=0.26\linewidth,height=0.22\linewidth]{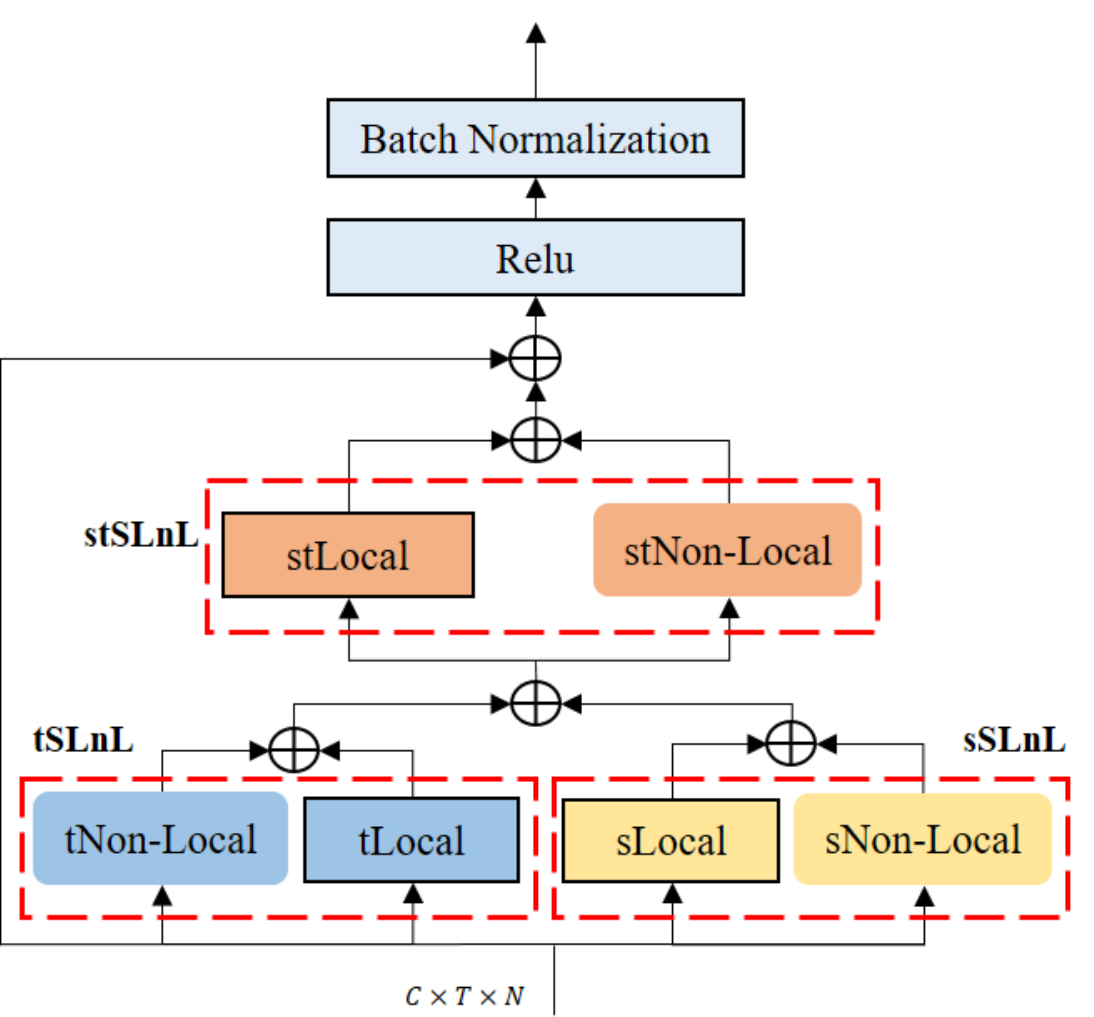}
	}
	\subfigure[The affinity field of SLnL]{
		\label{Fig3:d}
		\includegraphics[width=0.26\linewidth,height=0.22\linewidth]{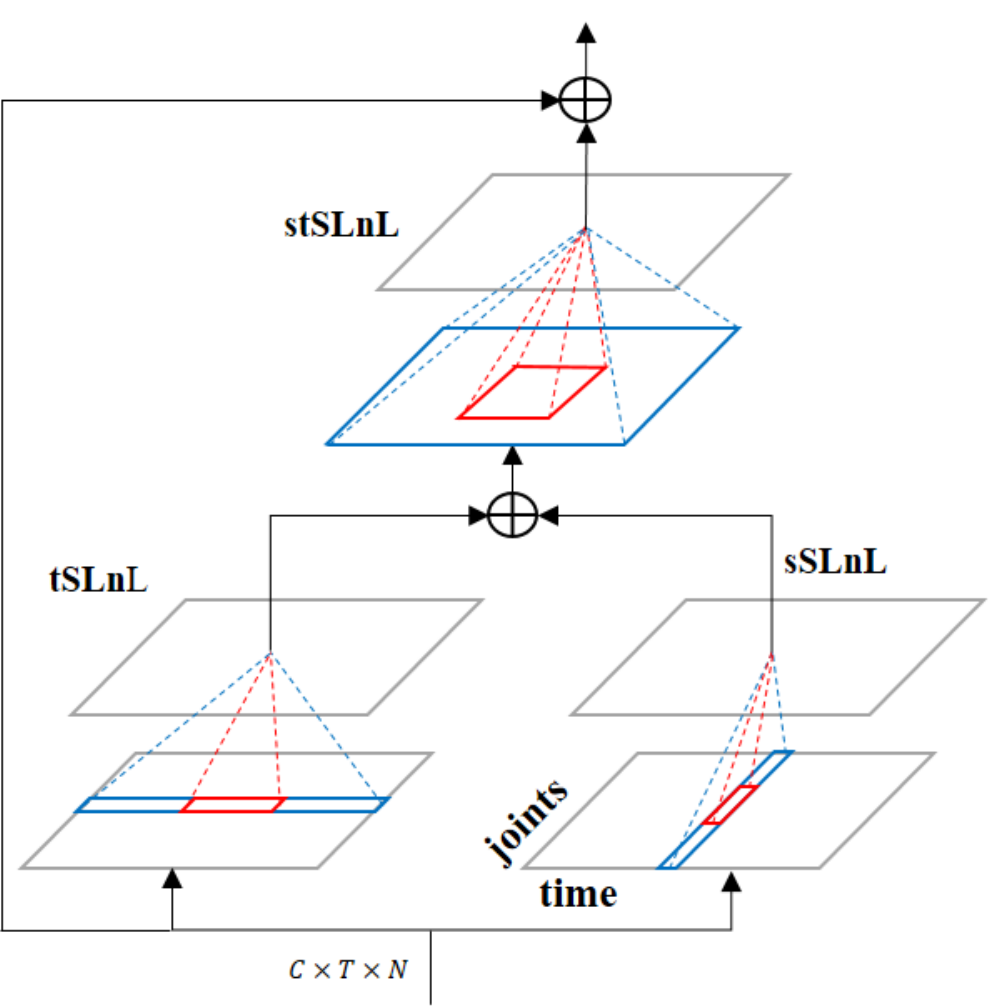}
	}
	\caption{(a) A 2D  example of non-local module. (b) The structure of the baseline local block. (c) The structure of the proposed synchronous local and non-local (SLnL) block. (d) The affinity field of SLnL. Note that the \textit{affinity field} is a more general concept than the \textit{receptive field} of CNNs. The red and blue represent local and non-local modules repectively in (d).}
	\label{Fig3}
\end{figure*}

Given a transformed action after the transform network {\myfont$\bm{X'}\in \mathbb{R}^{C' \times T' \times N'} $} ({\myfont $C'$=$Kd$}, $T'$=$T$ ), the 2D discret Fourier transform (DFT) transforms the pseudo spatio-temporal image {\myfont$\bm{X'}$} in each channel to
{\myfont$\bm{Y'} \in \mathbb{R}^{C' \times T' \times N'} $} in the
frequency domain via
{\myfont
	\begin{align}
	\bm{Y'}[c,u,v]
	& = \sum_{t=0}^{T'-1}\sum_{n=0}^{N'-1} \bm{X}'[c,t,n]cos\left(-2\pi\left(\frac{ut}{T'}+ \frac{vn}{N'}\right)\right) \notag\\
	& + j\sum_{t=0}^{T'-1}\sum_{n=0}^{N'-1} \bm{X}'[c,t,n]sin\left(-2\pi\left(\frac{ut}{T'}+ \frac{vn}{N'}\right)\right)  \notag\\
	& = \bm{F}_{cos}[c,t,n] + j \bm{F}_{sin}[c,t,n], \notag
	\end{align}
}where $u,v$ and $c$ are frequencies and channel of spatio-temporal image respectively, and {\myfont$\bm{F}_{cos}$}/{\myfont$ \bm{F}_{sin} $} denotes the cosine/sinusoidal component. The frequency spectrum {\myfont$\bm{F}_{A} = (\bm{F}_{cos}^2+ \bm{F}_{sin}^2)^{1/2}$} and the phase spectrum {\myfont$\bm{F}_{\phi} = arctan\left(-\frac{\bm{F}_{sin}}{\bm{F}_{cos}}\right) $}. In practice, the DFT and its inverse (IDFT) are computed through the fast Fourier transform (FFT) algorithm and its inverse (IFFT). 

For each action, the attention weights {\myfont$ \bm{M}_{cos} $} and {\myfont$ \bm{M}_{sin} $} are complex functions of its cosine and sinusoidal components, \ie~, 
{\myfont
	\begin{align}
	\bm{M}_i = dup(\sigma(\bm{W}_{i1}(\bm{W}_{i2}(Avg(\bm{F}_{i}))+b_{i1})+b_{i2})),
	\end{align}
}where {\myfont$ i \in \{cos,sin\} $}. Specifically, after a channel averaging operation, each component is fed into two fully connected layers (FC) to learn adaptive weights for each frequency, followed by a sigmoid transfom function. The first FC layers serve as a bottleneck layer \cite{DBLP:conf/cvpr/HeZRS16}  for dimensionality reduction with a ratio factor $ \lambda $. Then, the learned attention weights are duplicated to every channel to pay attention to the input frequency image via
{\myfont
	\begin{align}
	\bm{F}_{sin}'& = \bm{F}_{sin} \odot \bm{M}_{sin},\\
	\bm{F}_{cos}'& = \bm{F}_{cos} \odot \bm{M}_{cos},
	\end{align}
}where $ \odot $ denotes the element-wise multiplication. Finally, a spatio-temporal residual component is applied to obtain the output {\myfont$\bm{X''}\in \mathbb{R}^{C' \times T' \times N'} $} after attention, \ie~
{\myfont
	\begin{align}
	\bm{X}''& = \bm{X}'+ ifft\it{2} (\bm{F}_{sin}',\bm{F}_{cos}'),
	\end{align}
}where {\myfont$ ifft\it{2} $} denotes the efficient $ 2 $-dimensional IFFT.

\subsection{Synchronous Local and Non-local Learning in the Spatio-temporal Domain}
\textbf{Non-local Module.}
A general non-local operation takes a multi-channel signal {\myfont$ \bm{X}\in \mathbb{R}^{M \times P} $} as its input and generates a multi-channel output {\myfont$ \bm{Y}\in \mathbb{R}^{M \times Q} $}. Here {\myfont$ P $} and  {\myfont$ Q $} are channels, and {\myfont$ M $} is the number of {\myfont$ \Omega $}, where {\myfont$ \Omega $} is the set that enumerates all positions of the signal (image, video, feature map, etc.). Let {\myfont$ \bm{x_i} $} and {\myfont$ \bm{y_i} $} denote the $ i $-th row vector of  {\myfont$ \bm{X} $} and {\myfont$ \bm{Y} $}, 
the non-local operation is formulated as follows:
{\myfont
	\begin{align}
	\bm{y}_i = \frac{1}{\mathcal{Z}_i(\bm{X})}\sum_{j \in \Omega} \phi (\bm{x}_i,\bm{x}_j) g(\bm{x}_j), && \forall i \in \Omega
	\end{align}
}where the multi-channel unary transform {\myfont$ g(\bm{x}_j) $} computes the embedding of $ x_j $, the multi-channel binary transform {\myfont$ \phi (\bm{x}_i,\bm{x}_j) $} computes the affinity between the positions $ i $ and $ j $, and {\myfont$ \mathcal{Z}(\bm{X}) $} is a normalization factor. With different choices of $ \phi $ and $ g $, such as Guassian, embeddded Gaussian and dot product, various of non-local operations could be constructed. For simplicity, we only consider $ \phi $ and $ g $ in the form of linear embedding and embeddded Gaussian respectively, and set {\myfont$ \mathcal{Z}_i(\bm{X}) =\sum_{j\in \Omega}\phi (\bm{x}_i,\bm{x}_j) $}, \ie~
{\myfont
	\begin{align}
	g(\bm{x}_j)= (\bm{W}_g \bm{x}_j^T)^T, && \forall j
	\end{align}
}where {\myfont$ \bm{W}_g \in \mathbb{R}^{Q \times P} $} are learnable transform parameters.
{\myfont
	\begin{align}
	\phi (\bm{x}_i,\bm{x}_j) = e^{\varphi(\bm{x}_i)^T \psi(\bm{x}_j)},  && \forall i,j\\
	\varphi(\bm{x}_i) = (\bm{W}_{\varphi} \bm{x}_i^T)^T,  && \forall i\\
	\psi(\bm{x}_j) = (\bm{W}_{\psi} \bm{x}_j^T)^T,  && \forall j
	\end{align}
}where {\myfont$ \bm{W}_{\varphi},\bm{W}_{\psi} \in \mathbb{R}^{L \times P} $}, and $ L $ denotes the embedding channel. To weigh how important the non-local information is when compared to local information, a weighting function is appended,
\ie~
{\myfont
	\begin{align}
	w(\bm{y}_i) = (\bm{W}_w (\bm{y_i})^T)^T,
	\end{align}
}where {\myfont$ \bm{W}_{w} \in \mathbb{R}^{Q \times Q} $}. 
A non-local module 
can be completed with some transpose operations, some convolutional layers with the kernels of 1, and a softmax layer, Fig.\ref{Fig3:a} shows a 2D example.

\textbf{Baseline local block.}
The local operation is defined as
{\myfont
	\begin{align}
	\bm{y}_i = \frac{1}{\mathcal{Z}_i(\bm{X})}\sum_{j\in\delta_i} \phi (\bm{x}_i,\bm{x}_j)g(\bm{x}_j),  && \forall i \in \Omega
	\end{align}
}where {\myfont$ \delta_i $} is the local neighbor set of target position $  i $, {\myfont$ \delta_i\ll \Omega $}.
The convolution is a typical local operation with identity affinity {\myfont$ \phi (\bm{x}_i,\bm{x}_j) = 1 $}, liner transform {\myfont$ g(\bm{x}_j) = \bm{w}_j\bm{x}_j $}, identity normalization factor {\myfont$ \mathcal{Z}_i(\bm{X})=1$}, and {\myfont$ \delta_i $} is the neighbors around target center $ i $ with a same shape of kernel. Our baseline local block is constructed from convolution operation. As shown in Fig.\ref{Fig3:b}, two convolutional layers with kernel {\myfont$ k\times 1  $} and {\myfont$ 1\times k $} are applied to learn temporal local (tLocal) features and spatial local (sLocal) features respectively, and a {\myfont$ k \times k $} convolutional layer for spatial-temporal local (stLocal) features. 
The block also contains a residual path, a rectified linear unit (ReLU) and a batch normalization (BN) layer.

\textbf{Synchronous local and non-local block.} 
In order to synchronously exploit local details and non-local semantics in human actions, three non-local modules are parallel merged into the above baseline local block. As shown in Fig.\ref{Fig3:c}, two 1D non-local modules to explore temporal non-local (tNon-Local) and spatial non-local (sNon-Local) information respectively, followed by a 2D non-local module for spatio-temporal non-local (stNon-Local) patterns. We define the \textit{affinity field} as the representation of the range of pixel indices that could contribute to the target position in the next layer of the local or non-local modules, which is a more general concept than the \textit{receptive field} of CNNs. The affinity field in Fig.\ref{Fig3:d} clearly shows our SLnL can mine local details and non-local semantics synchronously in every layer. Note that our SLnL is significantly different from the methods \cite{DBLP:journals/corr/abs-1711-07971,DBLP:journals/corr/abs-1806-02919} which only inserted a few non-local modules after stacked local networks, thus the local and non-local operations are still separately conducted in different layers having different resolutions. Contrastively, our SLnL simultaneously captures local and non-local patterns in every layer (Fig.\ref{Fig3:d}).

\subsection{Soft-margin focal loss}
A common challenge for classification tasks is that the discrimination difficulties are different among samples and classes, but most previous works for skeleton-based action recognition use the \textit{softmax loss} that haven't taken it into consideration. There are two possible measures to alleviate it, \ie~data selection and margin encouraging.

\begin{figure}[tbp]
	\centering
	\includegraphics[width=0.95\linewidth,height=0.76\linewidth]{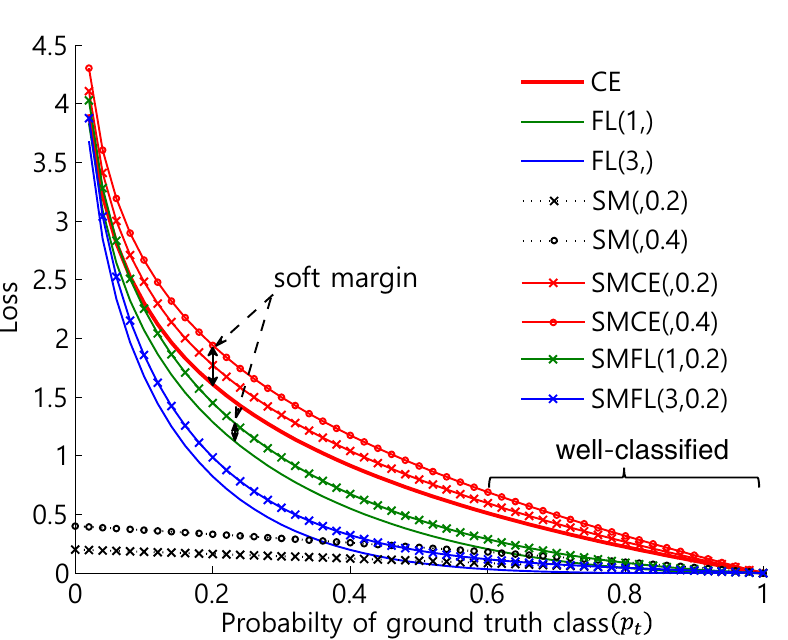}
	\caption{Comparisons among 
		soft-margin focal loss (SMFL), the soft-margin cross entropy (SMCE) loss, the cross-entropy (CE) loss, the focal loss (FL), and the soft-margin loss (SM). The focusing parameter $ \gamma $ and the margin parameter $ m $ of losses are expressed as $ (\gamma, m) $.}
	\label{fig4}
\end{figure}

Intuitively, the larger predicted probability a sample has, the farther away from the decision boundary it might be, and vice versa. Motivated by this intuition, we construct a soft-margin (SM) loss term as follows:
{\myfont
	\begin{align}
	\mathcal{L}_{SM}(p_t) = log\left( e^m+(1-e^m)p_t\right),
	\end{align}
}where $ p_t $ 
is the estimated posterior probability 
of ground truth class, and $ m $ is a margin parameter. {\myfont$ \mathcal{L}_{SM} \in [0,m]$} 
because that $ p_t\in[0,1] $. As Fig.\ref{fig4} shows when the posterior probability $ p_t $ is small, the sample is more likely be close to the boundary, thus we penalize it with a 
large margin loss. Otherwise, a small margin loss is imposed. To further illustrate the idea, we introduce the {\myfont$ \mathcal{L}_{SM} $} 
into 
cross entropy loss leading to a soft-margin cross entropy (SMCE) loss,
{\myfont
	\begin{align}
	\mathcal{L}_{SMCE}(p_t) & = \mathcal{L}_{SM} + \mathcal{L}_{CE} \\
	& = log\left( e^m+(1-e^m)p_t\right)-log(p_t). \notag
	\end{align}
}Assuming that {\myfont$ \bm{x}\in \mathbb{R}^d $} is the 
features before the last FC layer, the FC layer transforms it into score {\myfont$ \bm{z}= [z_1, z_2,\dots, z_C]^T \in \mathbb{R}^{C} $} of {\myfont$ C $} classes by multiplying {\myfont$ \bm{W} = [\bm{w}_1, \bm{w}_2, \cdots,\bm{w}_C] \in \mathbb{R}^{d \times C} $}, where {\myfont$ \bm{w}_c $} is the parameter of the linear classifier corresponding to the class $ c $, \ie~{\myfont$ z_c = \bm{w}_c^T \bm{x} $}. Followed with a softmax layer, 
{\myfont$ p_t = \frac{e^{\bm{w}_t\bm{x}}}{\sum_{c=1}^{C} e^{\bm{w}_c\bm{x}}}$} and {\myfont$ (1 - p_t) = \frac{\sum_{c\neq t}^{C}e^{\bm{w}_c\bm{x}}}{\sum_{c=1}^{C} e^{\bm{w}_c\bm{x}}}$}, then the SMCE can be rewritten as
{\myfont
	\begin{align}
	& \mathcal{L}_{SMCE}
	= log\left( p_t+e^m \cdot (1-p_t)\right)-log(p_t) \nonumber\\
	& = log\left( \frac{e^{\bm{w}_t\bm{x}}+e^m \cdot\sum_{c\neq t}^{C}e^{\bm{w}_c\bm{x}}}{\sum_{c=1}^{C} e^{\bm{w}_c\bm{x}}} \right) - log\left( \frac{e^{\bm{w}_t\bm{x}}}{\sum_{c=1}^{C} e^{\bm{w}_c\bm{x}}}\right) \notag\\
	& = -log \left( \frac{e^{\bm{w}_t\bm{x}}}{e^{\bm{w}_t\bm{x}}+e^m \cdot\sum_{c\neq t}^{C}e^{\bm{w}_c\bm{x}}}\right) \nonumber\\
	& = -log \left( \frac{e^{\bm{w}_t\bm{x}-m}}{e^{\bm{w}_t\bm{x}-m} + \sum_{c\neq t}^{C}e^{\bm{w}_c\bm{x}}}\right).
	\label{eq_smce}
	\end{align}
}Comparing the standard \textit{softmax loss} with Eq.\ref{eq_smce}, only the score of the ground truth class $ \bm{w}_t\bm{x} $ is replaced by $ \bm{w}_t\bm{x}-m $. Optimizing model with SMCE, we will obtain classifiers that meet the constraint {\myfont$ \bm{w}_t\bm{x}-m \geq \bm{w}_{c\neq t}\bm{x}  $}. As a result, an intrinsic margin $ m  $ between the positive (belonging to a specific class) samples and the negative (not belonging to the specific class) samples of each class will be formed in classifiers by adding the SM loss term into the loss function.

In addition, the focal loss \cite{DBLP:conf/iccv/LinGGHD17} defined as
{\myfont
	\begin{align}
	\mathcal{L}_{FL}(p_t) = -(1-p_t)^\gamma log(p_t),
	\end{align}
}where $ \gamma $ is a focusing parameter, can encourage adaptive data selection without any damage to the  original model structure and training processes. As Fig.\ref{fig4} shows the relative loss for well-classified easy samples is reduced by FL when compared to CE. Although FL pays more attention to hard samples, it has no margin around the decision boundary. Similar to SMCE, we introduce the {\myfont$ \mathcal{L}_{SM} $} term into FL to obtain the soft-margin focal loss (SMFL) as follows:
{\myfont
	\begin{align}
	\mathcal{L}_{SMFL}(p_t) & = \mathcal{L}_{SM} + \mathcal{L}_{FL}\\
	& = log\left( e^m+(1-e^m)p_t\right)-(1-p_t)^{\gamma}log(p_t).  \notag
	\end{align}
}Finally, our 
SMFL can encourage intrinsic margins in classifiers and maintain FL's advantage of 
data selection as well.

Our two stream model (Fig.\ref{fig_pipline}) predicts three probability vectors $ \bm{p}^p $, $ \bm{p}^v $, $ \bm{p}^c $\ from three modes including position, velocity, and their concatenation. We optimize it as a pseudo multi-task learning problem with our SMFL, \ie~each classifier produces a loss 
via
{\myfont
	\begin{align}
	\mathcal{L}_k = \sum_{i=1}^{C}y_i\left(log( e^m+(1-e^m)p_i^k)-(1-p_i^k)^{\gamma}log(p_i^k) \right),
	\end{align}
}where $ k\in \{p,v,c\}  $ is mode type, and {\myfont$ \bm{y}=(y_1,y_2,\cdots,y_C) $} is the one-hot class label. Thus the final loss is as follows:
{\myfont
	\begin{align}
	\mathcal{L} = \mathcal{L}_p + \mathcal{L}_v + \mathcal{L}_c.
	\end{align}
}During inference, only $ \bm{p}^c $ is used to predict the final class.

\section{Experiments}
\subsection{Datasets and Experimental details}
\textbf{NTU RGB+D (NTU)} dataset  \cite{DBLP:conf/cvpr/ShahroudyLNW16} is currently the largest in-door action recognition dataset. It contains 56,000 clips in 60 actions performed by 40 subjects. Each clip consists of 25 joint locations with one or two persons. There are two evaluation protocols for this dataset, \ie, cross-subject (CS) and cross-view (CV). For the cross-subject evaluation, 40320 samples from 20 subjects were used for training and 16540 samples from the rest subjects were used for testing. For the cross-view evaluation, samples are split by camera views, with two views for training and the rest one for testing.

\textbf{Kinetics} dataset 
is by far the largest unconstrained action recognition dataset, which contains 300,000 video clips
in 400 classes retrieved from YouTube \cite{DBLP:conf/aaai/YanXL18}. The skeleton 
is estimated by Yan \etal~from the raw RGB videos by OpenPose toolbox \cite{DBLP:conf/aaai/YanXL18}. Each joint consists of 2D coordinates $ (X,Y) $ in the pixel coordinate system and a confidence score $ C $, thus finally represented by a tuple of $(X,Y,C)$. Each skeleton frame is recorded as an array of 18 tuples. 

\textbf{Implementation Details:} During the data preparation,
we randomly crop sequences with a ratio uniformly drawn from [0.5,1] for training, and centrally crop sequences with a fixed ratio of 0.95 for inference. We resize the sequences to 64/128 (NTU/Kinetics) frame with bilinear interpolation. Finally, the obtained data are fed into a batch normalization layer to normalize the scale. During training, we apply Adam optimizer with  weight decay of 0.0005. Learning rate is initialized as 0.001, followed by an exponential decay with a rate of 0.98/0.95 (NTU/Kinetics) per epoch. A dropout with ratio of 0.2 is applied to each block to alleviate overfitting. The model is trained for 300/100 epoches with a batch size of 32/128 (NTU/Kinetics).

Each stream of model for NTU is composed of totally 6 blocks in Fig.\ref{Fig3} with local kernels of 3 and channels of 64, 64, 128, 128, 256, 256 respectively, also max-pooling is applied every two blocks. For Kinetics, two additional blocks with channels of 512 are appended, also the local kernels of the first two blocks are changed into 5. The numbers of new coordinate systems $ K $ and new joints $ N' $ in the transform network are set as 10 and 64 respectively for both datasets.

\begin{table}[tbp]
	\begin{minipage}[t]{0.49\linewidth}
		\caption{Comparisons of recognition accuracy (\%) on NTU.}
		\begin{threeparttable}
			\begin{tabularx}{1.0\linewidth}{l|>{\centering\arraybackslash}X>{\centering\arraybackslash}X}
				\hline
				Methods &  CS  & CV \\
				\hline
				PA-LSTM \cite{DBLP:conf/cvpr/ShahroudyLNW16} & 70.3 & 62.9 \\
				ST-LSTM+TG \cite{DBLP:conf/eccv/LiuSXW16}  & 69.2 & 77.7 \\
				VA-LSTM \cite{DBLP:conf/iccv/ZhangLXZXZ17} & 79.4 & 87.6 \\
				ST-GCN \cite{DBLP:conf/aaai/YanXL18}  & 81.5 & 88.3 \\
				TS-CNN \cite{DBLP:conf/icmcs/LiZXP17}  & 83.2 & 89.3 \\
				HCN \cite{DBLP:conf/ijcai/LiZXP18}  & 86.5 & 91.1 \\
				SR-TSL \cite{DBLP:journals/corr/abs-1805-02335} & 84.8 & 92.4 \\
				\hline
				SLnL-rFA (ours)  & {\bf 89.1} & {\bf 94.9} \\
				\hline
			\end{tabularx}
			\label{table_ntu}
		\end{threeparttable}
	\end{minipage}
	\hspace{0.02\linewidth}
	\begin{minipage}[t]{0.49\linewidth}
		\centering
		\caption{Comparing with the state-of-the-art approaches in action recognition accuracy (\%) on Kinetics dataset. Both of the top1 and top5 accuracies are reported. }
		\begin{threeparttable}
			\begin{tabularx}{1.0\linewidth}{l|>{\centering\arraybackslash}X>{\centering\arraybackslash}X}
				\hline
				Methods &  top1  & top5 \\
				\hline
				Feature Enc. \cite{DBLP:conf/cvpr/FernandoGMGT15}  & 14.9 & 25.8 \\
				Deep LSTM \cite{DBLP:conf/cvpr/ShahroudyLNW16}  & 16.4 & 35.3 \\
				Tem. Conv. \cite{DBLP:conf/cvpr/KimR17} & 20.3 & 40.0 \\
				ST-GCN \cite{DBLP:conf/aaai/YanXL18}  & 30.7 & 52.8 \\			
				\hline
				SLnL-rFA (ours)  & {\bf 36.6} & {\bf 59.1} \\
				\hline
			\end{tabularx}
			\label{table_kenetics}
		\end{threeparttable}				
	\end{minipage}
\end{table}

\subsection{Experimental Results}

On NTU RGB+D, we compare with three LSTM-based methods \cite{DBLP:conf/iccv/ZhangLXZXZ17,DBLP:conf/cvpr/ShahroudyLNW16,DBLP:conf/eccv/LiuSXW16}, two CNN-based methods \cite{DBLP:conf/ijcai/LiZXP18,DBLP:conf/icmcs/LiZXP17}, one graph convolutional method \cite{DBLP:conf/aaai/YanXL18}, and one graph and LSTM hybridized method \cite{DBLP:journals/corr/abs-1805-02335}. As the local components of our SLnL are CNN-based while the non-local components 
learn the affinity degree between each target position (node) to every position (node) in the figure (graph), our SLnL-rFA can be treated as a variant of CNN and graph hybridized method. As shown in Table \ref{table_ntu}, the CNN-based methods are generally better than LSTM-based methods, and graph-based or graph-hybridized methods also perform well. Our method consistently outperforms the state-of-the-art approaches by a large margin for both cross-subject (CS) and cross-view (CV) evaluation. Specifically, our SLnL-rFA outperforms the best CNN-based method (HCN) by 2.6\% (CS) and 3.8\% (CV), also outperforms the recent LSTM and graph hybridized method (SR-TSL) by 4.3\% (CS) and 2.5\% (CV).

On Kinetics, we compare with four characteristic methods, including hand-crafted features \cite{DBLP:conf/cvpr/FernandoGMGT15}, deep LSTM network \cite{DBLP:conf/cvpr/ShahroudyLNW16}, temporal convolutional network \cite{DBLP:conf/cvpr/KimR17}, and graph convolutional network \cite{DBLP:conf/aaai/YanXL18}. 
Table \ref{table_kenetics} shows the deep models outperform the hand-crafted features
, and the CNN-based methods work better than the LSTM-based methods. Our method outperforms the state-of-the-art approach (ST-GCN) by large margins of 5.9\% (top1) and 6.3\% (top5).

\subsection{Ablation Study}
To analyze the effectiveness of every proposed component, extensive ablation studies are conducted on NTU RGB+D.


\noindent\textbf{Comparisons on loss function.}
The baseline model (Baseline$_1$) of this section only contains local blocks in Fig.\ref{Fig3:b} and the transform network. The model is optimized with the cross entropy loss (CE), focal loss (FL), soft-margin cross entropy loss (SMCE), and soft-margin focal loss (SMFL), respectively. To save space, at most two best parameters for each loss are listed in Table \ref{table_loss}. Due to the adaptive data selection, FL performs better than CE. Benefiting from the encouraged margins between the positive and negative samples, 
the SMCE and SMFL perform better than their original versions CE and FL, respectively. Finally, our SMFL achieves the best for its advantages from adaptive data selection and intrinsic margin encouraging.

\noindent\textbf{How to select discriminative frequency patterns?} We firstly reform the Baseline$_1 $ into Baseline$_2 $ (No FA) for this section by adding the SMFL. To validate the effectiveness of proposed rFA, we compare it with several variants. The Amplitude frequency attention (aFA) is built on frequency spectrum instead of sinusoidal and cosine components. Shared FA (sFA) learns shared 
parameters for sinusoidal and cosine components, while dependent FA (dFA) learns two set of parameters independently. The rfA is formed by applying the residual learning trick to dFA in the spatio-temporal domain (Fig.\ref{fig2}). In Table \ref{table-fa}, we observe that aFA is harmful because the phase angle information is 
missing when only using the frequency spectrum. The dFA outperforms the sFA because that it has more parameters to model the frequency patterns. The rFA finally achieves the best that outperforms Baseline$_2$ with a large margin, indicating that the frequency information is effective for action recognition.

\begin{table}[hbp]
	\begin{minipage}[t]{0.46\linewidth}
		\caption{Results of different loss functions in accuracy (\%).}
		\begin{threeparttable}
			\begin{tabularx}{1.0\linewidth}{l|>{\centering\arraybackslash}X>{\centering\arraybackslash}X}
				\hline
				Loss types  & CS & CV \\
				\hline
				CE (Baseline$_1$)   & 85.5 & 91.3 \\
				FL{\small(2,)}  & 85.8 & 91.9 \\
				FL{\small(3,)} & 85.6 & 91.8 \\
				\hline
				SMCE{\small(,0.4)} & 86.4 & 92.0 \\
				SMCE{\small(,0.6)} & 86.2 & 92.3 \\
				SMFL{\small(2,0.4)} & {\bf 86.9} & 92.5 \\
				SMFL{\small(2,0.6)} &  86.5 & {\bf 92.6} \\ 			
				\hline	
			\end{tabularx}
			\label{table_loss}
		\end{threeparttable}
	\end{minipage}
	\hspace{0.01\linewidth}
	\begin{minipage}[t]{0.53\linewidth}
		\caption{Performance comparisons of different frequency attention methods in human action recognition accuracy (\%).}
		\begin{threeparttable}
			\begin{tabularx}{1.0\linewidth}{l|>{\centering\arraybackslash}X>{\centering\arraybackslash}X}
				\hline
				Attention methods &  CS  & CV   \\
				\hline
				No FA (Baseline$_2$) & 86.9 & 92.6 \\
				\hline
				Amplitude FA  & 84.7 & 89.8 \\
				Shared FA    & 87.3 & 92.9 \\
				Dependent FA  & 87.5 & 93.2 \\		
				Residual FA (rFA)\tnote  & {\bf 87.7} & {\bf 93.6} \\ 	
				\hline
			\end{tabularx}
		\end{threeparttable}	
		\label{table-fa}			
	\end{minipage}
\end{table}

\noindent\textbf{Comparisons of methods with different affinity fields.} We further reform the Baseline$_2 $ into Baseline$_3 $ with a rFA block for this section.
Although non-local dependencies can be captured in higher layers of hierarchical local networks, we argue that synchronously explore and fuse non-local information in early stages is preferable. We merge one temporal non-local block (tSLnL), spatial non-local block (sSLnL), or spatial-temporal block (SLnL) into Baseline$ _3 $ to examine their effectiveness. As shown in Table \ref{table-SLnL}, both the non-local information from the temporal and spatial dimensions during early stages are helpful. In addition, benefiting from the synchronous fusion of local details and non-local semantics, our SLnL boosts up the recognition performance by 1.4\% (CS) and 1.1\% (CV).
To further investigate the properties of deeper SLnL, we replace $ M_1 $ local blocks in Baseline$ _3 $ with SLnL. Table \ref{table-SLnL} shows more SLnL blocks in lower layers generally lead to better results, but the improvements of higher layers is relatively small because the affinity field of local operations is increasing with layers. The results clearly show that synchronously extracting local details and non-local semantics is vital for modeling the spatio-temporal dynamics of human actions.

\begin{table}[htbp]
	\centering
	\begin{threeparttable}
		\caption{Comparisons of methods with various affinity fields. $ M_1 $ and $ M_2 $ denotes the number of SLnL and local blocks in Fig.\ref{fig_pipline}, respectively.}
		\begin{tabularx}{0.88\linewidth}{l|>{\centering\arraybackslash}X|>{\centering\arraybackslash}X}
			\hline
			Affinity Field  &  CS (\%)  & CV (\%)  \\
			\hline
			Local (Baseline$_3$) & 87.7 & 93.6 \\
			\hline
			tSLnL ($ M_1 $ = 1, $ M_2 $ = 5) & 88.1 & 93.9 \\
			sSLnL ($ M_1 $ = 1, $ M_2 $ = 5) & 88.0 & 94.1 \\
			SLnL ($ M_1 $ = 1, $ M_2 $ = 5) & 88.3 & 94.3 \\
			\hline
			SLnL ($ M_1 $ = 2, $ M_2 $ = 4) & 88.6 & 94.6 \\
			SLnL ($ M_1 $ = 3, $ M_2 $ = 3) & 88.8 & {\bf 94.9} \\
			SLnL ($ M_1 $ = 4, $ M_2 $ = 2) & 88.9 & 94.8 \\
			SLnL ($ M_1 $ = 5, $ M_2 $ = 1) & {\bf 89.1} & 94.7 \\
			SLnL ($ M_1 $ = 6, $ M_2 $ = 0) & 88.8 & 94.7 \\
			\hline		
		\end{tabularx}
		\label{table-SLnL}
	\end{threeparttable}
\end{table}

\section{Conclusion}
In this work, we propose a novel model SLnL-rFA to extract synchronous detailed and semantic information from multi-domains for skeleton-based action recognition. The SLnL synchronously extracts local details and non-local semantics in the spatio-temporal domain. The rFA adaptively selects discriminative frequency patterns, which sheds a new light to exploit information in the frequency domain for skeleton-based action recognition. In addition, we also propose a novel soft-margin focal loss, which can encourage intrinsic margins in classifiers and  conducts adaptive data selection.  Our approach significantly outperforms other state-of-the-art methods both on the largest in-door dataset NTU RGB+D and on the largest unconstrained dataset Kinetics for skeleton-based action recognition.

\bibliographystyle{IEEEbib}
\bibliography{icme2019template,biblist}

\end{document}